\documentclass[runningheads]{llncs}

\usepackage{eccv}

\usepackage{eccvabbrv}

\usepackage{graphicx}
\usepackage{booktabs}
\usepackage{color, colortbl}
\usepackage{stackengine}
\usepackage[dvipsnames]{xcolor}

\usepackage[accsupp]{axessibility}  % Improves PDF readability for those with disabilities.

\usepackage{hyperref}

\usepackage{orcidlink}
\definecolor{lightyellow}{HTML}{FFF2CC}
\definecolor{lightviolet}{HTML}{E1D5E7}
\definecolor{lightblue}{HTML}{DAE8FC}

\definecolor{darkyellow}{HTML}{D6B656}
\definecolor{darkviolet}{HTML}{9673A6}
\definecolor{darkblue}{HTML}{6C8EBF}

\begin{document}

% \title{Trans\textit{V}Geo: Transformer is All You Need for Cross-view \textit{V}ideo Geo-localization} 

\title{GAReT: Cross-view Video \underline{G}eolocalization with \underline{A}dapters and Auto-\underline{Re}gressive \underline{T}ransformers}
% \title{TransCVGL: Transformer-based Cross-view Video Geo-localization}

%is All You Need for Cross-view \textit{V}ideo Geo-localization}

\titlerunning{GAReT}
% \titlerunning{TransCVGL}

\author{Manu S Pillai\inst{1} \and
Mamshad Nayeem Rizve\inst{2} \and
Mubarak Shah\inst{1}}

\authorrunning{M.~Pillai et al.}

\institute{Center for Research in Computer Vision, University of Central Florida, USA \and
Adobe\\
\email{manu.pillai@ucf.edu}, 
\email{mrizve@adobe.com}
\email{shah@crcv.ucf.edu}\\}

\maketitle
\begin{abstract}
Cross-view video geo-localization (CVGL) aims to derive GPS trajectories from street-view videos by aligning them with aerial-view images. Despite their promising performance, current CVGL methods face significant challenges.
These methods use camera and odometry data, typically absent in real-world scenarios. They utilize multiple adjacent frames and various encoders for feature extraction, resulting in high computational costs. Moreover, these approaches independently predict each street-view frame's location, resulting in temporally inconsistent GPS trajectories.
To address these challenges, in this work, we propose {\textbf{GAReT}}, a fully transformer-based method for CVGL that does not require camera and odometry data. We introduce \textit{GeoAdapter}, a transformer-adapter module designed to efficiently aggregate image-level representations and adapt them for video inputs. Specifically, we train a transformer encoder on video frames and aerial images, then freeze the encoder to optimize the \textit{GeoAdapter} module to obtain video-level representation.
To address temporally inconsistent trajectories, we introduce \textit{TransRetriever}, an encoder-decoder transformer model that predicts GPS locations of street-view frames by encoding top-$k$ nearest neighbor predictions per frame and auto-regressively decoding the best neighbor based on the previous frame's predictions. Our method's effectiveness is validated through extensive experiments, demonstrating state-of-the-art performance on benchmark datasets. Our code is available at \href{https://github.com/manupillai308/GAReT}{\texttt{https://github.com/manupillai308/GAReT}}.

\end{abstract}
% \vspace{-0.3in}
    
\section{Introduction}
\label{sec:intro}
Recently, there has been significant interest in cross-view video geo-localization (CVGL), driven by its applications in vision-guided navigation, autonomous driving, and robot manipulation~\cite{berrabah2011visual, berton2022rethinking, brosh2019accurate}. Like its image counterpart~\cite{shi2019spatial,zhang2023geodtr+,geodetr,zhu2021vigor,zhu2022transgeo}, the objective of CVGL is to obtain the GPS trajectory of a video by matching every frame of a query street-view video with a reference aerial-view gallery. 
Prior art primarily tackled CVGL from a coarse Sequence-to-Image~\cite{shi2022cvlnet,zhang2023cross} matching perspective. In this setting, the objective is to obtain a \textit{single} GPS coordinate by matching a sequence of street-view images (typically $\sim 7$)~\cite{zhang2023cross}, sampled from a street-view video with geo-tagged aerial images in the reference gallery. However, this coarse localization may not be optimal for many real-world applications~\cite{brejcha2017state}. Hence,~\cite{vyas2022gama} extends CVGL to a more fine-grained Frame-to-Frame setting, where each frame of a street-view video is matched with a corresponding reference aerial image (frame-level association). Particularly,~\cite{vyas2022gama} proposes a hierarchical approach and a hierarchical dataset, where they first localize a street-view video to a larger geographical area before obtaining the final frame-by-frame predictions. Thus, following~\cite{vyas2022gama,regmi2021video,shi2022cvlnet,zhang2023cross}, tackling CVGL at both coarse (sequence-to-image) and fine (frame-to-frame) levels is the primary focus of this work. 

While existing works on CVGL have demonstrated promising performance, they are not without notable drawbacks. Firstly, these approaches often rely on camera intrinsic parameters and odometry data to model the temporal relation in the street-view video and to effectively match it with a reference aerial image~\cite{shi2022cvlnet}, which are not always available, especially in videos captured in uncontrolled environments. Secondly, to improve performance, prior work~\cite{vyas2022gama} utilizes contextual information from multiple adjacent street-view frames (0.5-second short `clip') to obtain a feature representation for the center street-view frame. This design leads to additional computational overhead for obtaining per-frame features and, hence, is not optimal for real-time applications. Lastly, and most importantly, existing works on CVGL suffer from temporally inconsistent GPS predictions due to modeling per-frame localization as independent nearest-neighbor retrieval operations~\cite{vyas2022gama,shi2022cvlnet}. However, while obtaining the GPS predictions for each frame in a street-view video, the predictions should follow temporal closeness consistent with the input video. In this work, we aim to address these limitations of existing CVGL approaches. 

To this end, we extend TransGeo~\cite{zhu2022transgeo}, a versatile transformer-based method for cross-view image geo-localization, to the video setting and propose \textbf{GAReT}, a fully %completely 
{trans}former-based method for {c}ross-view {v}ideo {g}eo-{l}ocalization. We hypothesize that in the context of video geo-localization, the temporal reasoning requirement, though important, is not as critical as in other video tasks (like action recognition). Following this assumption, we hypothesize that in order to geo-localize a street-view video, the representations from a model capable of geo-localizing street-view images can be efficiently aggregated to obtain video-level representation. Towards this, we propose a transformer-adapter-based strategy (\textit{GeoAdapter}) to aggregate image-level representations of an image geo-localization method and adapt it to video inputs. The advantage of this strategy is two-fold: (i) As the image geo-localization method is grounded on matching frame-level representations, a learnable aggregation module can effectively fuse the representations to obtain representation for video geo-localization. (ii) The efficiency of adapters allows our model to be more computationally lightweight than traditional video models~\cite{bertasius2021space,liu2022video} as well as previous works~\cite{vyas2022gama,shi2022cvlnet}. Additionally, our method tackles the problem of CVGL in two hierarchical stages (as opposed to four in~\cite{vyas2022gama}), where we first localize a street-view video to a large aerial region (using the video representation) and then individually localize each frame of the video within the larger region. As transformers can model view-specific feature attributes without explicit view transformations (e.g., polar transform or projective transform)~\cite{zhu2022transgeo}, our method does not require any knowledge of camera intrinsic or odometry. Furthermore, to ensure temporal consistency in frame-to-frame predictions, we propose an encoder-decoder transformer model (\textit{TransRetriever}) that encodes the top-$k$ nearest neighbor predictions for each frame and auto-regressively decodes the best result based on previous frames.

We summarize our contributions as follows:
\begin{itemize}
    \item We propose a fully {trans}former-based method (\textbf{GAReT}) for {c}ross-view {v}ideo {g}eo-localization that doesn't rely on explicit viewpoint information like odometry and camera intrinsic. 
    \item We employ a novel transformer adapter (\textit{GeoAdapter}) for CVGL that adapts an image geo-localization encoder for video inputs, outperforming dedicated video models on benchmark datasets.
    \item To ensure temporally consistent frame-to-frame GPS predictions, we present the first learnable transformer-based model (\textit{TransRetriever}) that models the independent frame retrievals through an auto-regressive decoder.
\end{itemize}
% \vspace{-0.2in}
\section{Related Works}
Cross-view geo-localization represents an active field of research in computer vision with various applications in real-world systems. Diverse endeavors have been undertaken in the realm of cross-view \textit{image} geo-localization with limited efforts in its \textit{video} counterpart. 
\textbf{Cross-view Image geolocalization:} In cross-view image geo-localization, the aim is to obtain a GPS coordinate for a street-view image by matching it with aerial reference images.
In \cite{shi2019spatial}, a polar transform of aerial images was used to align views, alongside the introduction of the Spatial-Aware Feature Aggregation (SAFA) module for learning embeddings with spatial correspondence between views. To align views, \cite{liu2019lending} proposed using an orientation mask that provides geometric orientation correspondence to the feature extractor.
In \cite{shi2020looking, shi2022accurate}, the authors introduce Cross-DSM, a dynamic similarity matching (DSM) module to compute a rough estimate of the azimuth angle before performing matching.
\cite{yang2021cross} showed polar transform of aerial images is not necessarily required for better performance and introduced a transformer encoder with self-cross attention layers.
\cite{zhu2021revisiting} identified key factors in cross-view image geo-localization, especially the need for orientation alignment of street-view and aerial-view image pairs.
VIGOR~\cite{zhu2021vigor} dataset was introduced to address the lack of challenging benchmark datasets.
Following this,~\cite{zhu2022transgeo} proposed the first pure transformer-based architecture that performed significantly better than CNN counterparts without explicit view-specific transformation.~\cite{geometric_layout} propose a novel disentangled feature learning framework that outperforms other methods on cross-area benchmarks. Recently, a new direction in cross-view image geo-localization has emerged. The goal is to accurately determine the location of the street-view camera and estimate its pose~\cite{fervers2023uncertainty,beyond,lentsch2023slicematch,xia2022visual,xia2023convolutional}. 

As previously mentioned, the research landscape for \textit{video} geo-localization is considerably limited. \textbf{Cross-view Video geolocalization:}~\cite{shi2022cvlnet} developed a cross-view video-based localization approach using street-and-aerial geometric correspondences.
Another work~\cite{zhang2023cross} proposes a novel temporal feature aggregation technique to learn better global representations of street-view video. Though promising results were obtained for both these works, they tackled the problem of CVGL from a Sequence-to-Image standpoint.~\cite{vyas2022gama} introduced the most recent work in Frame-to-Frame CVGL, inspired by~\cite{regmi2021video}.
Their contribution entailed a pioneering hierarchical approach for cross-view video geo-localization where the query street-view videos are first localized to a larger aerial region, then localizing each video frame within the larger region. They employ different models for each localization stage, resulting in increased computational complexity.
Additionally, existing approaches in CVGL adhere to the nearest-neighbor paradigm for inference, where the best match from the reference set (aerial images) is taken as the nearest neighbor of the query (street-view video frame). This greedy nearest-neighbor strategy yields a suboptimal local solution by independently retrieving matches for rather temporally related frames. Several works have shown improvement over this standard local optimum by posing a global constraint or gain on the retrieval task \cite{zamir2014image, tian2017cross,zamir2014visual}, but all of them are heuristic-based approaches. 
In this work, our primary focus centers on mitigating these issues. We demonstrate how a purely transformer-based approach, equipped with our proposed \textit{GeoAdapter} module, can surpass current state-of-the-art methods by a significant margin. Furthermore, to ensure a temporally consistent solution, we propose the first transformer-based approach, \textit{TransRetriver}, that models independent frame retrievals through an auto-regressive transformer decoder.

% \vspace{-0.1in}
\section{Methodology}
\label{sec:method}
We first present the problem formulation and an overview of our approach in~\cref{sec:prob_form}. Then, we introduce the different components of our method in detail in~\cref{sec:img_repr}  and~\cref{sec:geoadapter}. Finally, we present \textit{TransRetriever}, our novel transformer-based learnable architecture for temporally consistent frame-to-frame predictions in~\cref{sec:transretri}.

% \vspace{-0.1in}
\subsection{Problem Formulation \& Approach}
\label{sec:prob_form}
We focus on CVGL in a frame-to-frame setting. Given a \textit{query} street-view video, we need to geo-localize each frame individually by visually matching the frames to geotagged aerial \textit{reference} images. Formally, given a set of query street-view videos $\{V^s\}$ and aerial-view reference images $\{I^a\}$, our objective is to learn an embedding space where every frame of query street-view videos is close to its corresponding ground-truth aerial image. 
Following the frame-to-frame setting proposed in~\cite{vyas2022gama}, we assume that each street-view video $V^{s_i}$ is associated with a \textit{large} aerial image in addition to aerial images (\textit{small}) corresponding to each street-view video frame. 
The large aerial image facilitates the hierarchical approach of localizing an entire street-view video to a larger geographical region (sequence-to-image) before individually obtaining the predictions for each frame (frame-to-frame). Precisely, each video $V^{s_i}$ of $n$ frames will have $n$ ground-truth small aerial images (denoted hereon by $\{I^{a_i}_{s_j}\}_{j=1}^n$) and one ground-truth large aerial image (denoted hereon by $I^{a_i}_L$).

\begin{figure}[h]
% \vspace{-6mm}
\centering
\includegraphics[width=0.9\textwidth]{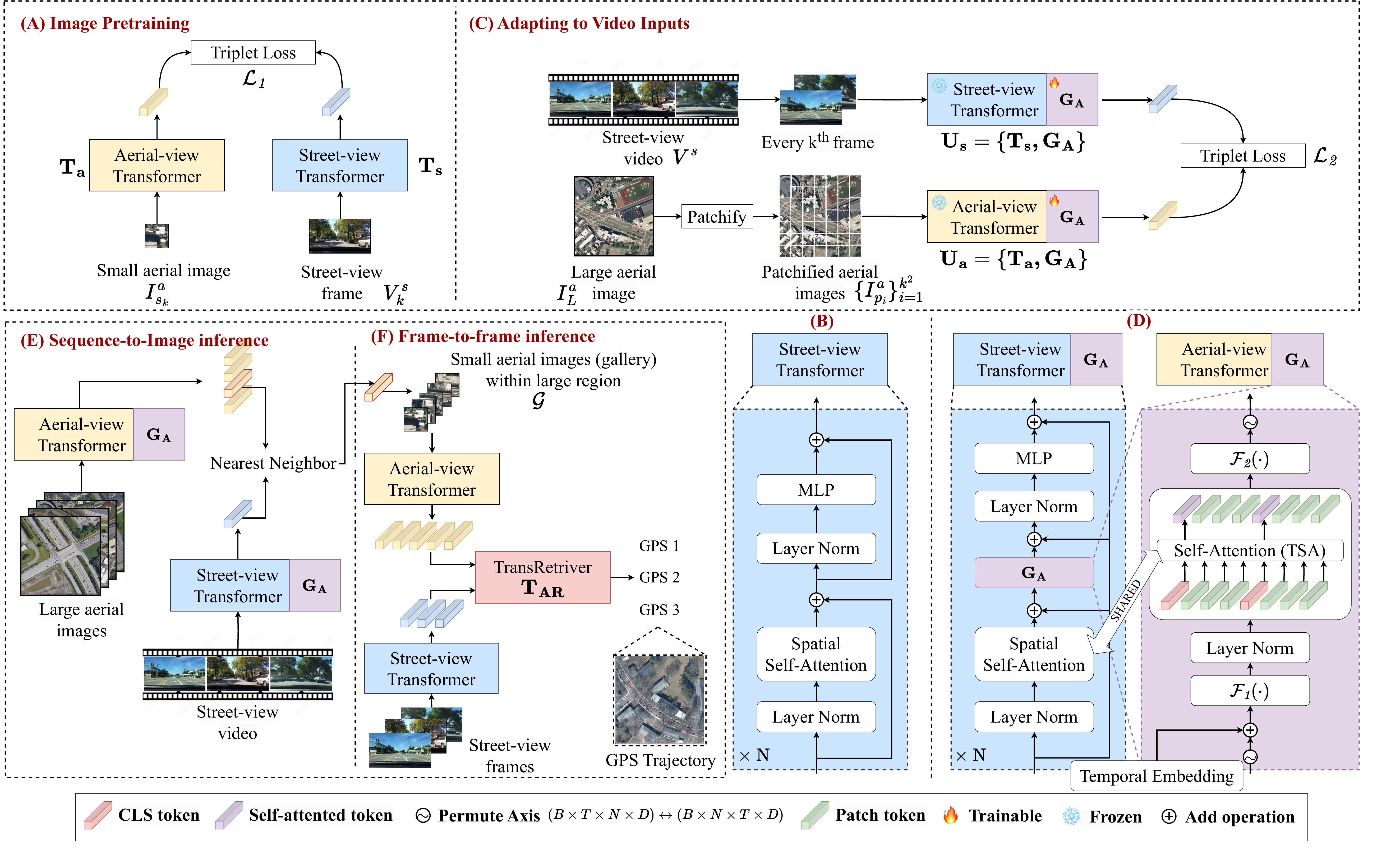}
\caption{Overview of our proposed approach GAReT. \textbf{(A)} We begin by optimizing our image transformer encoders $\mathbf{T_{a,s}}$ \textbf{(B)} with street-view frame $V^s_k$ and matching small aerial image $I_{s_k}^a$ pair. \textbf{(C)} Then, for adapting our image encoder to video inputs, we add our GeoAdapter $\mathbf{G_A}$ module and only optimize the adapter parameters with video pairs as inputs, i.e., a street-view video $V^s$ and corresponding large aerial image $I^a_L$. For training, we sample every $k^\text{th}$ frame from the street-view video and partition the large aerial image into non-overlapping patches.
\textbf{(D)} In $\mathbf{G_A}$, we apply temporal self-attention (TSA) computation only on the \texttt{CLS} tokens. For TSA computation, we reuse the spatial self-attention weights. \textbf{(E)} During inference, we first perform a Sequence-to-Image inference procedure, where given a query street-view video, our unified module $\mathbf{U = \{T, G_A\}}$ produces feature embeddings for both the $V^s$ and $I^a_L$. Then, using the embeddings, we retrieve the $t$ nearest neighbor large aerial images (here we show $t=1$) and construct a small aerial image gallery $\mathcal{G}$. \textbf{(F)} Finally, $\mathbf{G_A}$ is removed, and feature embeddings for $I_{s_k}^a$ and $V^s_k$ are obtained. These features are then passed to our TransRetriever $\mathbf{T_{AR}}$ model to obtain final frame-by-frame GPS predictions to construct a GPS trajectory.}
% \vspace{-0.1in}
\label{fig:main}
\end{figure}

As shown in~\cref{fig:main}, we propose a hierarchical solution towards CVGL. During training, we optimize our feature encoder with image and video inputs to learn a unified model ($\mathbf{U}$) for feature extraction. Our unified model is a combination of a vision transformer (ViT) ($\mathbf{T}$) as in~\cite{zhu2022transgeo} equipped with our proposed GeoAdapter ($\mathbf{G_A}$) module, i.e., $\mathbf{U} = \{\mathbf{T}, \mathbf{G_A}\}$. We begin by optimizing the ViT parameters with image pairs (street-view frame and small aerial image) as inputs while keeping the adapter module frozen. Then, we take video pairs (street-view video and large aerial image) as inputs and freeze the transformer encoder parameters, and optimize the adapter module. 
During inference, we localize a given street-view video to a larger region using our unified model and then independently obtain frame-to-frame predictions using the vision transformer (without the proposed GeoAdapter) encoder within the localized larger region. Finally, instead of independently retrieving frame-to-frame predictions through nearest-neighbor retrieval, we use our proposed TransRetriever ($\mathbf{T_{AR}}$) to autoregressively predict GPS coordinates for each frame by selecting matching small aerial images from a set of possible candidates (see~\cref{sec:transretri}). 

% \vspace{-0.15in}
\subsection{Image Representations for CVGL}
\label{sec:img_repr}
Following~\cite{zhu2022transgeo}, we use DeiT~\cite{pmlr-v139-touvron21a} as image encoder for our unified model $\mathbf{U}$. Note that, for simplicity, we consider one pair of street-view and aerial images ($V^{s_i}, I^{a_i}$) and ignore subscripts in notations. 

Given a pair $(V^s, I^a)$, we sample every $k^\text{th}$ second frame from $V^s$ and obtain corresponding matching $I_{s_k}^a \in \mathbb{R}^{256\times256}$. For each view-specific input, similar to~\cite{shi2019spatial,zhang2023geodtr+,geodetr,zhu2021vigor,zhu2022transgeo}, we train two separate image encoders, i.e., $\mathbf{T_a, T_s}$ to generate embedding features. Specifically, we divide an input image ($V^s_k$ or $I_{s_k}^a$) of size $H, W$ into $P$ non-overlapping patches of size $p\times p$. Each patch is flattened and fed into a trainable linear projection layer to generate token representations. The tokens are prepended with a learnable token representation called \texttt{[CLS]} token to learn global features from the image. Each token is then added with positional embeddings~\cite{vaswani2017attention} and goes through multiple transformer blocks. The encoded \texttt{[CLS]} token representation after the last transformer block is passed through a linear projection layer to obtain the final embedding of the input image (additional details are provided in the supplementary material).

The encoders are optimized through contrastive learning~\cite{radford2021learning} to facilitate the model in learning embeddings where matching pairs are encoded closer together while non-matching pairs are encoded farther apart. Formally, we optimize the encoders on soft-margin triplet loss~\cite{hu2018cvm} as given below:
\begin{equation}
    \mathcal{L}_1 = \sum_{V^s}\sum_{k}\log{(1+e^{\alpha(||\mathbf{T_s}(V^s_k) - \mathbf{T_a}(I_{s_k}^a)||_2 - ||\mathbf{T_s}(V^s_k) - \mathbf{T_a}(I_{s_{i\neq k}}^a)||_2)})}
\end{equation}
where, $||\cdot||_2$ denotes $l2$ norm and $\mathbf{T_{a,s}}: \mathbb{R}^{H\times W\times 3} \rightarrow \mathbb{R}^{d}$. 

% \vspace{-0.15in}
\subsection{GeoAdapter: Adapting to Video for CVGL}
\label{sec:geoadapter}
In order to localize a street-view video to a larger geographical region, we extend our image geolocalization model $\mathbf{T_{a,s}}$ to video inputs. We propose GeoAdapter ($\mathbf{G_A}$), the second and final component of our unified model $\mathbf{U}$ to efficiently aggregate the representations of $\mathbf{T_{a,s}}$ to obtain video embeddings. $\mathbf{G_A}$ is a transformer adapter module inspired by~\cite{yang2022aim,pan2022st} which consists of two 2-layer MLPs with GeLU activation function, separated by a temporal self-attention (TSA) layer. The TSA layer models the temporal relation between frames of the input video by leveraging self-attention across the temporal dimension of the input. 
Furthermore, to induce temporal ordering in tokens, we add temporal positional embeddings to each of the tokens before computing TSA. To effectively learn rich contextual information from the video in different levels, we add our proposed module to every block of the transformer encoder $\mathbf{T_{a,s}}$. Formally, in block $l$ of our unified model $\mathbf{U} = \{\mathbf{T}, \mathbf{G_A}\}$, we have, 
\begin{equation}
    \begin{split}
        \hat{h}^l &= h^{l-1} + \mathrm{SA}(\mathrm{LN}(h^{l-1}))\\
        {h}^l &= \hat{h}^l + \mathbf{G_A}(\hat{h}^l, \mathrm{TE})\\
        h^l &= {h}^l + \mathrm{MLP}(\mathrm{LN}({h}^l))\\
    \end{split}
\end{equation}
where $\mathbf{G_A}(\hat{h}^l, \mathrm{TE}) = \mathcal{F}_2(\mathrm{TSA}(\mathrm{LN}(\mathcal{F}_1(\hat{h}^l + \mathrm{TE}))))$, $h^{l-1}$ is output of block $l-1$, $\mathrm{SA, LN}$ and $\mathrm{TE}$ is self-attention, layer-norm and temporal positional embedding respectively. $\mathcal{F}_{1,2}$ is 2-layer MLPs with GeLU activation function. Not that for TSA computation, we reuse the spatial self-attention weights similar to~\cite{yang2022aim, pan2022st}.\\
\textbf{\underline{Discussion}.} As compared to~\cite{yang2022aim,pan2022st}, our adapter module has two major differences. (1) As the image transformer encoder $\mathbf{T_{a,s}}$ is trained using frames of the video, % that we extend to (using adapter), 
we don't need to perform any spatial adaptation. (2) As discussed in~\cref{sec:intro}, the temporal modeling for CVGL can be greatly relaxed due to the simplicity of the task. Particularly, to model temporal relations, we do not need to compute temporal attention with every patch of every input frame. Instead, we can use the global representation of every frame (\texttt{[CLS]} token) and attend them across the temporal dimension. To facilitate this, we compute the spatial self-attention prior to computing TSA in every block, which is in reverse order to~\cite{yang2022aim}. We experimentally demonstrate that, for CVGL, computing TSA on \texttt{[CLS]} tokens outperforms TSA on all patches in terms of retrieval performance and computational efficiency (see~\cref{sec:add_expr}).
% \vspace{-0.2in}
\subsubsection{Adapting the aerial encoder $\mathbf{T_a}$:}\label{sec:adapt_aerial} In order to extend the aerial encoder $\mathbf{T_a}$ to obtain embeddings for larger geographical regions, we perform a pre-processing step before adding $\mathbf{G_A}$. %Recall that, 
% each \textit{large} aerial image covers approximately 49 times more area than a \textit{small} aerial image. This implies that at the same zoom level [footnote], 
%a large aerial image is 7 times the size of a small aerial image (see~\cref{sec:prob_form}). 
Since each \textit{large} aerial image covers approximately $49$ times (for GAMa dataset~\cite{vyas2022gama}) more area than a \textit{small} aerial image, to ensure that there is no distribution shift in the encoder $\mathbf{T_a}$, which has been trained on small aerial images, we divide the large aerial image into $49$ non-overlapping patches. Each patch is of the same size %(thus the same Ground Sampling Distance [cite]) 
as the small aerial image. Formally, given $I_L^a \in \mathbb{R}^{kn\times kn}$ with $I^a_{s}\in \mathbb{R}^{n\times n}$, we break $I^a_L$ into $k^2$ non-overlapping patches of size $n\times n$, i.e., $I^a_L = \{I^a_{p_i}\}_{i=1}^{k^2}$. These patches of images can be considered analogous to the street-view frames but for the aerial encoder. Note that there is no one-to-one correspondence between the aerial patch $I^a_{p_i}$ and street-view frame $V^s_k$. We only know matching pairs of street-view videos and large aerial images.

Similar to~\cref{sec:img_repr}, the unified model $\mathbf{U}$ is trained using contrastive learning with soft-margin triplet loss. We freeze the encoders $\mathbf{T_{a,s}}$ and optimize only the added GeoAdapter modules $\mathbf{G_A}$. Given street-view videos and large aerial images, our unified model learns to embed matching pairs closer to each other while pushing non-matching pairs farther apart. Formally, we optimize the following objective:
\begin{equation}
    \mathcal{L}_2 = \sum_{i}\log{(1+e^{\alpha(||\mathbf{U_s}(V^{s_i}) - \mathbf{U_a}(I^{a_i}_{L})||_2 - ||\mathbf{U_s}(V^{s_i}) - \mathbf{U_a}(I^{a_{k\neq i}}_{L})||_2)})}
    \label{eq:l2}
\end{equation}
where, $||\cdot||_2$ denotes $l2$ norm and $\mathbf{U_{a,s}}: \mathbb{R}^{N\times H\times W\times 3} \rightarrow \mathbb{R}^{d}$.\\ 
% \textbf{\underline{Discussion}.} Due to the minimal distribution shift between the inputs used to train each component of the unified model, we empirically found that we could train both the models $\mathbf{T}$ and $\mathbf{G_A}$ alternatively one after the other without loss in performance. Specifically, given a street-view video $V_s$, matching small aerial images $\{I^{a}_{s_j}\}_{j=1}^n$ and large aerial image $I^a_L$, we can train the unified model $\mathbf{U}$ by optimizing the following objective:
% \begin{equation}
% \begin{split}
%     \mathcal{L} = \lambda\cdot \mathcal{L}_1 + (1-\lambda)\cdot \mathcal{L}_2\\
%     \lambda = 
%     \begin{cases}
%     1, & \mbox{if } \mbox{epoch }\%\ K = 0 \\ 
%     0, & \mbox{otherwise} 
%     \end{cases}
% \end{split}
% \end{equation}
% where $\%$ is the remainder operator, and $K$ is a hyperparameter that specifies the interval to alternate training between $\mathbf{G_A}$ and $\mathbf{T}$. 
% \vspace{-0.25in}
\subsection{TransRetriever: Temporally Consistent Retrieval}
\label{sec:transretri}
% For a given street-view video $V_s$, we can obtain $k$ large aerial regions $\{I^{a_k}_L\}$ using our unified model $\mathbf{U}$. Then, we can perform frame-by-frame matching (within the $k$ regions) using $\mathbf{T}$ to obtain final GPS predictions for each frame. Specifically, 
% Given $V_s$, we obtain the $k$ nearest neighbor large aerial images in the embedding space of $\mathbf{U}$. Using the obtained $k$ large regions, we create a small aerial reference gallery $\mathcal{G}$, similar to [cite]. For each frame $V^s_i$ of the street-view video, we can obtain the nearest neighbor small aerial images in the embedding space of $\mathbf{T}$ and predict the GPS coordinates. This nearest-neighbor strategy yields temporally inconsistent GPS predictions as shown in~\cref{fig:temp}. Towards this, our proposed TransRetriever ($\mathbf{T_{AR}}$) aims to autoregressively predict GPS coordinates by selecting a matching small aerial image from a set of possible candidates. Precisely, given
Given $V_s$, we obtain the $k$ nearest neighbor large aerial images in the embedding space of $\mathbf{U}$. Using the obtained $k$ large images, we create a small aerial reference gallery $\mathcal{G}$, similar to~\cite{vyas2022gama}. For each frame $V^s_i$ of the street-view video, we obtain $t$ nearest neighbor small aerial images in the embedding space of $\mathbf{T}$ from $\mathcal{G}$. In order to obtain final temporally consistent frame-to-frame GPS predictions, we formulate the retrieval task as the %in the realm of 
Generalized Traveling Salesman problem (GTSP)~\cite{gtsp}. Precisely, given $V^s = \{V^s_1, V^s_2, ..., V^s_n\}$ with ${N} = \{{N}_1, {N}_2, ..., {N}_n\}$ where $N_i=\{I^{a}_{s_1}, I^{a}_{s_2}, ..., I^{a}_{s_t}\}$ is the set of $t$ small aerial nearest neighbor of $V^s_i$, we want to predict one small aerial image from each set $N_i$ such that, the prediction from $N_i$ and $N_{i+1}$ is in closer proximity to each other.
% \textbf{\underline{Discussion}.} If we consider each small aerial image as a city, each set $N_i$ is a cluster of cities. From the perspective of GTSP, the task is to find the shortest path (or time) to visit every cluster of cities exactly once. The distance (or time taken) from each city to another city can be formed as a combination of the GPS distance of each city (GPS distance between small aerial images $I^{a}_{s_i}$ and $I^{a}_{s_j}$) and the visual similarity of each city in a cluster with its street-view frame ($\mathbf{T_s}(V^s_i)^T\cdot\mathbf{T_a}(I^{a}_{s_j})$ where $I^{a}_{s_j} \in N_i$).
% \vspace{-0.2in}
\subsubsection{Overview:} %The architectural design of the proposed TransRetriever is inspired by [cite, cite]. 
Similar to~\cite{bresson2021transformer}, we cast our GTSP problem as a `translation' problem where the source `language' is the set $N$ (analogous to clusters of cities), and the target `language' is $n$ small aerial images, one from each $N_i$ (analogous to one city from each cluster). Analogous to translation architectures in NLP~\cite{cho2014properties}, our proposed TransRetriever has an encoder-decoder architecture, i.e., $\mathbf{T_{AR}} = \{\mathbf{E}, \mathbf{D}\}$. The source `language', i.e., the set $N$, is encoded by the encoder, and the decoder autoregressively predicts the target small aerial images using the encoded representations. Since the elements of the set $N$ are ordered (as the frames of $V^s$ are ordered), we use positional encoding to induce the ordering before encoding. To obtain a global context before decoding, we append a learnable token \texttt{[START]} (analogous to \texttt{[CLS]} token in VIT~\cite{dosovitskiy2020image}) to the set $N$. The decoder takes the encoder representation of the \texttt{[START]} token to begin the decoding phase, where at every iteration $i$, we obtain a small aerial image that is the final prediction for street-view frame $V^s_i$. A comprehensive architecture diagram for $\mathbf{T_{AR}}$ is provided in the supplementary material. 
% \vspace{-0.2in}
\subsubsection{Methodology:} %As shown in~\cref{fig:main}, 
We begin by obtaining a 3-dimensional representation for each small aerial in $N$. Precisely, for all $N_i \in N$, for all $I^a_{s_j} \in N_i$, we convert its GPS coordinate to UTM coordinates and obtain its similarity score $S_{ij} = C_S(\mathbf{T_a}(I^a_{s_j}), \mathbf{T_s}(V^s_i))$, where $C_S$ is cosine similarity. Each small aerial image $I^a_{s_j} \in N_i$ is then represented by a 3-D vector $r_{ij} = [\mathrm{UTM_x}, \mathrm{UTM_y}, S_{ij}]$. We will refer to $r_{ij}$ as tokens from hereon for simplicity. Since each $N_i$ represents a set of neighbors for street-view frame $V^s_i$, an inherent ordering exists for all $N_i$ and $N_{i+1}$.  To induce such an ordering in the encoding phase, we add positional embedding to each $r_{ij}$ based on the $N_i$ it belongs to. Specifically, we add a positional embedding $\mathrm{PE}_i$ to every $r_{ij}$ in $N_i$. Note that all the tokens belonging to one neighbor set $N_i$ will be added with the same positional embedding $\mathrm{PE}_i$, e.g., $r_{12}$ and $r_{13}$ will be added with $\mathrm{PE}_1$ while $r_{23}$ and $r_{21}$ will be added with $\mathrm{PE}_2$ and so on. The resulting tokens $x_{ij} = r_{ij} + \mathrm{PE}_i$ along with the learnable token \texttt{[START]} is encoded through the encoder $\mathbf{E(\cdot)}$ to obtain $[h_{\text{\texttt{[START]}}}, h_{ij}] = \mathbf{E}([{\text{\texttt{[START]}}}, x_{ij}])$. 

Next, we utilize the encoded representation $h_{\text{\texttt{[START]}}}$ as a starting token for the decoder to predict the small aerial image for the first street-view frame $V^s_1$. Similar to the encoding phase, to induce ordering as well as to enforce position-specific decoding, we add the encoded representation with its corresponding positional embedding before passing it to the decoder. Formally, in the decoder $\mathbf{D(\cdot)}$, to predict the small aerial image for the $i^\text{th}$ street-view frame $V^s_i$, we add the encoded token $h_{(i-1)j}$ (the prediction of the previous frame $V^s_{i-1}$) with positional embedding $\mathrm{PE}_i$ and pass it to the decoder and compute causal self-attention (masked self-attention)~\cite{vaswani2017attention,bresson2021transformer} to induce autoregressive conditioning followed by a cross-attention with only the encoded representations $r_{ij} \in N_i$. This enforces that the prediction of the $i^\text{th}$ street-view frame comes only from set $N_i$. The decoding starts with $h_{0j} = h_{\text{\texttt{[START]}}}$. For each input token $h_{ij}$, the decoder predicts a probability distribution $p_{i}$ that represents the probability of each encoded token in $N_i$ to be selected. During inference, we take the token with the highest probability as prediction and pass it to the next iteration of the decoder. Formally, we have,
$p_{i} = \mathbf{D}(h_{(i-1)j} + \mathrm{PE}_i, [h_{i1}, h_{i2}, ..., h_{it}])$ with $\hat{j}_i = \text{argmax}(p_{i})$ and $h_{(i+1)j} = h_{i{\hat{j}_i}}$. Evidently, the prediction for the $i^\text{th}$ street-view frame $V^s_i$ will be $r_{i{\hat{j}_i}}$ which we convert back to GPS coordinates to obtain the final GPS predictions.

\vspace{-0.15in}
\section{Experiments and Results}
\label{sec:exp}
\vspace{-0.05in}
\subsection{Training and Implementation details}
We use DeiT-m~\cite{pmlr-v139-touvron21a} as our image transformer architecture and use PyTorch~\cite{paszke2019pytorch} to implement our framework. We train the models $\mathbf{T}$ and $\mathbf{G_A}$ for 100 epochs with a batch size of 64 and 8, respectively. For both models, we use Adam~\cite{kingma2014adam} optimizer with a learning rate of 1e-4 and default settings for other parameters. We use pre-trained weights from ImageNet to initialize our image encoder $\mathbf{T}$ and initialize all the parameters in the $\mathbf{G_A}$ module with 0. For our proposed TransRetriever architecture $\mathbf{T_{AR}}$, we have a 6-layer encoder with a 2-layer decoder with embedding dimension 128 and feed-forward dimension 512. Following~\cite{bresson2021transformer}, we use Batch Normalization~\cite{ioffe2015batch} instead of Layer Normalization~\cite{ba2016layer}. We use sinusoidal positional embedding~\cite{vaswani2017attention} and train the model for 500 epochs.
% \newpage

\vspace{-0.1in}
\subsection{Datasets}
\vspace{-0.05in}
\subsubsection{GAMa:} GAMa~\cite{vyas2022gama} dataset is a cross-view video geo-localization (CVGL) dataset that aims to provide a comprehensive benchmark for evaluating CVGL methods in a frame-to-frame setting. The datasets provide $\sim 40$ seconds long street-view videos from the BDD100K~\cite{yu2020bdd100k} dataset with matching aerial images corresponding to each geo-tagged street-view frame of the video. Additionally, each video is accompanied by a large aerial image that helps to localize the entire video to a larger geographical region before obtaining frame-by-frame predictions. The dataset provides 45029 street-view videos and large aerial image pairs with 1.68 million frame-level matching aerial images.
\vspace{-0.1in}
\subsubsection{SeqGeo:} SeqGeo~\cite{zhang2023cross} dataset is an image sequence geolocalization dataset consisting of high-resolution frontal camera street-view images of $1920\times1080$ resolution with a field-of-view of approximately $120^\circ$. Each street-view image is a part of a video shot at 1 FPS, with the distance between each frame (or capture point) at approximately 8 meters. Additionally, each street-view image accompanies a GPS location and camera heading (compass direction) angle. The total number of street-view images is 118,549, grouped into sequences of $\sim 7$ images covering 50 meters of ground distance per sequence. Each sequence includes an aerial image of the region covered by the street-view images. In total, there are 38,863 pairs of aerial images and street-view image sequences covering approximately 500 kilometers of roads in Vermont.

% \subsection{Large Aerial Inference}
\vspace{-0.1in}
\subsection{Sequence-to-Image Inference}
\label{exp:large_aerial}
To assess the effectiveness of our approach, we initially showcase the performance of our method in retrieving large aerial images given a street-view video. We use our trained unified model $\mathbf{U}$ to encode the given street-view video and find the best matching large aerial image by calculating the nearest neighbor in the feature space. Following previous works~\cite{shi2022cvlnet,zhang2023cross,vyas2022gama}, we report top-k retrieval recall at $k=1$, 5, 10, and 1\% to compare our model with other state-of-the-art methods. A match is correct if the retrieved $k$ nearest neighbours consists the correct large aerial image. We compare our proposed approach on the GAMa~\cite{vyas2022gama} dataset against Screening Network from~\cite{vyas2022gama}, CVLNet~\cite{vyas2022gama} (since CVLNet requires ground-truth camera parameters, we compare our method with the "Ours w/o GVP (Unet)" and "Ours w/o GVP" variants for fairness.), and general-purpose video models VideoSWIN~\cite{liu2022video} (initialized with pretrained weights of Kinetics400~\cite{kay2017kinetics}), TimeSformer~\cite{bertasius2021space}(initialized with pretrained weights of ImageNet~\cite{deng2009imagenet}). To further establish the effectiveness of our GeoAdapter module $\mathbf{G_A}$, we compare our method with two additional baselines that are extensions of TransGeo~\cite{zhu2022transgeo} to video inputs.\\
\textbf{Baseline 1:} As the first baseline, we train the image encoder $\mathbf{T_{a,s}}$ with video inputs without our proposed adapter $\mathbf{G_A}$. Specifically, we perform the pre-processing step on the large aerial image $I^a_L$ (see~\cref{sec:adapt_aerial}) and encode each patch individually using the encoder $\mathbf{T_a}$. The embedding of the large aerial image is thus the average of the individual embeddings of each patch, $I^a_{p_i}$. Precisely, $f_L = \frac{1}{k^2}\sum_{i=1}^{k^2}\mathbf{T_a}(I^a_{p_i})$, where $k^2$ is the total number of non-overlapping patches of the large aerial image $I^a_L$. Similarly, the embedding for a street-view video $V^s$ is obtained by computing the average of the individual embeddings of each frame $V^s_i$, i.e., $f_V = \frac{1}{n}\sum_{i=1}^\mathbf{T_s}(V^s_i)$, where $n$ is the total number of frames in $V^s$. We train the encoders $\mathbf{T_{a,s}}$ with~\cref{eq:l2} with $f_V$ and $f_L$ as video and large aerial embeddings. Note that, for a fair comparison, we initialize the encoders $\mathbf{T_{a,s}}$ with weights obtained after pretraining with the images following~\cref{sec:img_repr}.\\
\textbf{Baseline 2:} To establish the effectiveness of our preprocessing step during adaptation of $\mathbf{T_a}$ with our GeoAdapter module (see~\cref{sec:adapt_aerial}), we compare our approach with another baseline. In this, we perform the adaptation of only the street-view encoder $\mathbf{T_s}$ and train $\mathbf{T_a}$ (without the adapter) by resizing the large aerial image to size $512\times512$. Additional details about these experiments can be found in supplementary material.\\
% \textbf{Baseline 2 vs Our approach:} Our approach uses GeoAdapter to adapt the aerial encoder with the large aerial image preprocessed into a sequence of images, while baseline-2 resizes the larger aerial image to a lower resolution and treats it as an image for the encoder.
\vspace{-0.02in}
~\cref{tab:vgl} shows the quantitative result of our approach on large aerial inference. The baseline 1 performs better than both general-purpose video models, establishing the advantage of image-level pretraining in CVGL. Also, our method with both $\mathbf{T}$ and $\mathbf{G_A}$ performs superior to all other methods, achieving a top-1 recall rate of 50.69. To further demonstrate the performance, we compare our method against the state-of-the-art cross-view sequence geo-localization benchmark SeqGeo~\cite{zhang2023cross}. Note that even though SeqGeo is an image sequence dataset (sequence-to-image), our method can be applied here without loss of generality. Specifically, instead of $\sim 40$ street-view images (as in GAMa), SeqGeo consists of $\sim 7$ street-view images per aerial image. As it can be inferred from the results that, our method also performs significantly better on the SeqGeo benchmark nearly doubling the state-of-the-art achieving a top-1 recall rate of 3.34.\\
\textbf{\underline{Discussion}.} It is important to note that while Baseline 1 excels in large aerial inference, the encoders used to achieve this are entirely trained on video data. Therefore, once the larger aerial region is localized, the same encoder cannot be used for frame-by-frame matching. Whereas, with our proposed approach, once the large aerial region is obtained, we can disengage the adapter $\mathbf{G_A}$ in the encoder and use it for frame-by-frame matching. Further details on this behavior can be found in the supplementary material.
% \vspace{-0.2in}
\begin{table}[h]
    % \vspace{-0.25in}
    \centering
    \setlength{\tabcolsep}{0.6em} % for the horizontal padding
    {\renewcommand{\arraystretch}{1.1}
    \resizebox{30em}{!}{\begin{tabular}{lcccc}
    % \hline
    \toprule
    Model & R@1 $(\uparrow)$ & R@5 $(\uparrow)$ & R@10 $(\uparrow)$ & R@1\% $(\uparrow)$\\
    % \hline\hline
    \toprule
    \multicolumn{5}{c}{GAMa dataset} \\
    % \hline\hline
    \hline
    TimeSformer~\cite{bertasius2021space} & 20.10 & 44.47 & 55.58 & 83.46 \\
    % \hline
    VideoSWIN~\cite{liu2022video} & 20.37 & 45.86 & 59.94 & 88.02\\
    % \hline\hline
    CVL-Net~\cite{shi2022cvlnet} & 0.36 & 1.33 & 2.66 & 15.35 \\
    % \hline
    GAMa(Screening-Network)~\cite{vyas2022gama} & 12.2 & - & 35.3 & 49.3 \\
    % \hline\hline
    Baseline 1 & 40.07 & 70.76 & 79.72 & 94.04\\%1814400
    % \hline
    Baseline 2 & 15.39 & 38.48 & 51.36 & 83.16\\%1814400
    % \hline
    \rowcolor{lightblue}\textbf{Ours} $\mathbf{(T + G_A)}$ & \textbf{50.69} & \textbf{81.77} & \textbf{88.71} & \textbf{98.26}\\%1814400
    % \hline
    % \hline
    \toprule
    \multicolumn{5}{c}{SeqGeo dataset} \\
    \hline
    VIGOR~\cite{zhu2021vigor} & 0.54 & 2.52 & 4.48 & 18.55\\
    % \hline
    SAFA~\cite{shi2019spatial} & 0.63 & 2.83 & 5.03 & 21.51 \\
    % \hline
    SeqGeo~\cite{zhang2023cross}& 1.80 & 6.45 & 10.36 & 34.38\\
    % \hline
    \rowcolor{lightblue}\textbf{Ours} $\mathbf{(T + G_A)}$ & \textbf{3.34} & \textbf{11.19} & \textbf{17.18} & \textbf{44.39} \\
    % \hline
    \toprule
    \end{tabular}
    }}
    \vspace{0.05in}
    \caption{Top-k retrieval recall score for large aerial image localization. Our approach ($\mathbf{T} + \mathbf{G_A}$), achieves superior top-1 recall performance of 50.69 on the GAMa dataset, surpassing both baseline models and state-of-the-art methods. Additionally, our method significantly improves upon the state-of-the-art SeqGeo benchmark, achieving a top-1 recall rate of 3.34.
    (Arrows represent more $(\uparrow)$ or less $(\downarrow)$ is better, `-' denotes scores not provided in respective works)}
    \label{tab:vgl}
\vspace{-0.2in}
\end{table}
% \vspace{-0.7in}
\begin{figure}
    \centering
    \includegraphics[width=\textwidth]{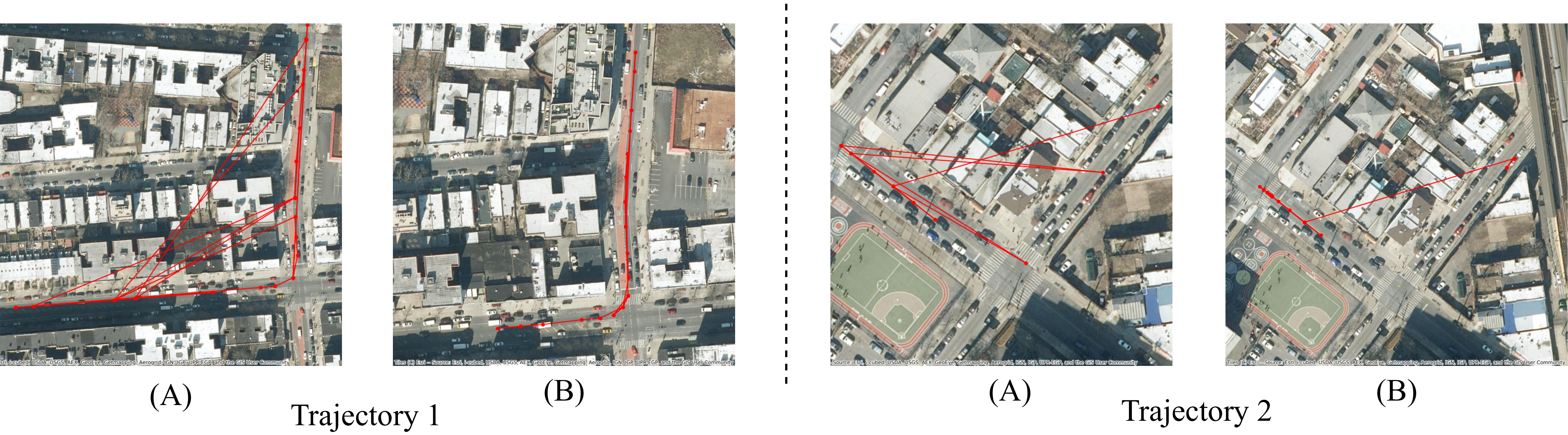}
    \caption{Examples of trajectories obtained using NN \textbf{(A)} based retrieval and our proposed TransRetriever \textbf{(B)}. NN-based retrieval heavenly suffers from temporally inconsistent predictions depicted by the jumps in the trajectory while TransRetriever predictions are globally consistent which preserves the temporal coherence of the predictions.}% \caption{Figure caption}
    \label{fig_transretri}
\vspace{-0.2in}
\end{figure}
\vspace{-0.2in}
\subsection{Frame-to-frame inference}
\label{sec:frame_inf}
After localizing a given street-view video to a large aerial region, similar to~\cite{vyas2022gama}, we create a small aerial image reference gallery $\mathcal{G}$. To obtain predictions for each street-view frame of the input video, we use our image encoder model $\mathbf{T_{a,s}}$ to retrieve nearest neighbor small aerial images from the gallery $\mathcal{G}$. Similar to~\cref{exp:large_aerial}, we report top-$k$ retrieval recall at $k=1$, 5, 10 and 1\%. Due to high degree of overlap in aerial images of subsequent video frames, following the previous work~\cite{vyas2022gama}, we consider a match correct if the predicted GPS is within the range of 0.05 miles of the ground-truth location. We compare our framework against GAMa-Net~\cite{vyas2022gama}, L2LTR~\cite{yang2021cross}, VIGOR~\cite{zhu2021vigor} and~\cite{shi2019spatial}.~\cref{tab:fgl} shows the quantitative results of our method in comparison to other methods. Note that the recall rate for GAMa-Net uses top-1\% large aerial images for creating the small aerial reference gallery, while our method uses top-10 large aerial images. Our proposed framework achieves significant improvement over state-of-art scoring 54.64\% top-1 recall rate on the validation split of the GAMa dataset. 
\vspace{-0.2in}
\subsection{Retrieval with TransRetriever}
% \vspace{-0.1in}
Following~\cref{sec:transretri}, we compare our proposed TransRetriever with nearest neighbor (NN) retrieval and Dominant sets (DS) strategy from [cite]. For predicting using NN, we obtain the nearest small aerial image to a given street-view frame individually and predict its GPS coordinates as the final prediction. For DS and TransRetriever, we obtain the $k$ nearest small aerial image candidates for each frame and predict the GPS coordinates by selecting one small aerial image from each $k$ neighbor (see~\cref{sec:transretri}). Additional information on how we implement the DS strategy for our use case can be found in the supplementary material. In~\cref{tab:trans_retri}, we show the top-1 recall rate of our method compared to NN and DS for $k=10$ and $k=20$.~\cref{fig_transretri} shows examples of the GPS trajectory obtained for a street-view video using NN and TransRetriever. The \textcolor{red}{red} point represents the GPS prediction for a frame in the video, and lines connect adjacent frame predictions. NN-based retrieval heavenly suffers from temporally inconsistent predictions depicted by the jumps in the trajectory. At the same time, TransRetriever predictions are globally consistent, reducing inconsistencies and preserving the temporal coherence in the predictions.
\begin{table}
\vspace{-0.2in}
\begin{minipage}[b]{.48\linewidth}
    \centering
    \resizebox{15em}{!}{\begin{tabular}{lc|c}
    % \hline
    \toprule
    Method & \multicolumn{2}{c}{R@1 $(\uparrow)$} \\
    % \hline\hline
    \hline
    & $k=10$ & $k=20$\\
    \hline
    Nearest Neighbor & \multicolumn{2}{c}{54.64}\\
    Dominant Sets~\cite{tian2017cross} & 56.09 
    & 60.67\\
    % \hline
    \rowcolor{lightblue}TransRetriever $(\mathbf{T_{AR}})$ & \textbf{67.66} & \textbf{70.70}\\
    % \hline
    \toprule
    \end{tabular}
    }
    \vspace{0.05in}
    \caption{The top-1 frame-by-frame recall rate of our method with nearest neighbor, dominant sets, and our proposed TransRetriever with $k=10$ and $k=20$. (Symbols follow the definition in~\cref{tab:vgl})}
    \label{tab:trans_retri}
\end{minipage}\hfill\begin{minipage}[b]{.49\linewidth}
    \centering
    {\renewcommand{\arraystretch}{1.1}
    \resizebox{18em}{!}{\begin{tabular}{lccc}
    % \hline
    \toprule
    Adapter & R@1 $(\uparrow)$ & R@5 $(\uparrow)$ & R@10 $(\uparrow)$ \\
    % \hline\hline
    \hline

    AIM~\cite{yang2022aim} &  48.06 &  80.50 & 88.25 \\

    ST-Adapter~\cite{pan2022st} & 34.29 & 69.98 & 80.75\\
    % \hline
    \rowcolor{lightblue}\textbf{Ours} & \textbf{50.69} & \textbf{81.77} & \textbf{88.71}\\
    % \hline
    \toprule
    \end{tabular}}
    }
    \vspace{0.05in}
    \captionof{table}{Top-$k$ retrieval recall score of Sequence-to-Image inference with AIM~\cite{yang2022aim}, ST-Adapter~\cite{pan2022st}, and our proposed GeoAdapter (Ours) as adapter modules. (Symbols follow the definition in~\cref{tab:vgl})}
    \label{tab:comp_adapt}
\end{minipage}
\vspace{-0.2in}
\end{table}
\vspace{-0.25in}

\begin{table} % for the horizontal padding
% -----------------------------------
    \vspace{-0.2in}
    \centering
    \setlength{\tabcolsep}{0.6em} % for the horizontal padding
    {\renewcommand{\arraystretch}{1.1}
    \resizebox{30em}{!}{\begin{tabular}{lcccc}
    % \hline
    \toprule
    Model & R@1 $(\uparrow)$ & R@5 $(\uparrow)$ & R@10 $(\uparrow)$ & R@1\% $(\uparrow)$ \\
    % \hline\hline
    \hline
    Shi et al.~\cite{shi2020looking} & 9.6 & 18.1 & 26.6 & 71.9\\
    % \hline
    L2LTR~\cite{yang2021cross} & 11.7 & 20.8 & 28.2 & 87.1\\
    % \hline
    GAMa-Net~\cite{vyas2022gama} & 15.2 & 27.2 & 33.8 & 91.9 \\
    % \hline
    GAMa-Net (Hierarchical)~\cite{vyas2022gama}  & 18.3 & 27.6 & 32.7 & - \\
    \rowcolor{lightblue}\textbf{Ours} ($\mathbf{T}$) & \textbf{54.64} & \textbf{70.45} & \textbf{76.36} & \textbf{91.92}\\
    % \hline
    \toprule
    \end{tabular}
    }}
    \vspace{0.05in}
    \caption{Top-k retrieval recall score for frame-by-frame geo-localization on the GAMa dataset. For our method, we construct the gallery $\mathcal{G}$ using the top 10 large aerial images, while other methods use the top 1\% images. Compared to the state-of-the-art, our method achieves significant improvement achieving a final score of 54.65 on frame-to-frame cross-view geo-localization. (Symbols follow the definition in~\cref{tab:vgl})}
    \label{tab:fgl}
    \vspace{-0.5in}
\end{table}
\vspace{-0.05in}
\subsection{Analysis}
\begin{figure}[htp]
\vspace{-0.4em}
\begin{minipage}[t]{.48\linewidth}
\bottominset{%
\resizebox{6.5em}{!}{\begin{tabular}{l|c}
    \hline
    Variant & R@1 \\
    \hline\hline
    \textcolor{darkblue}{$\mathbf{G_A^\text{ALL}}$}  & 48.4 \\
    \hline
    \textcolor{darkyellow}{$\mathbf{G_A^\text{ASYM}}$}  & 50.0\\
    \hline
    \textcolor{darkviolet}{\textbf{Ours} ($\mathbf{G_A}$)}  & \textbf{50.7}\\
    \hline
    \end{tabular}}
}{%
\scalebox{1.}{\includegraphics[width=0.9\columnwidth]{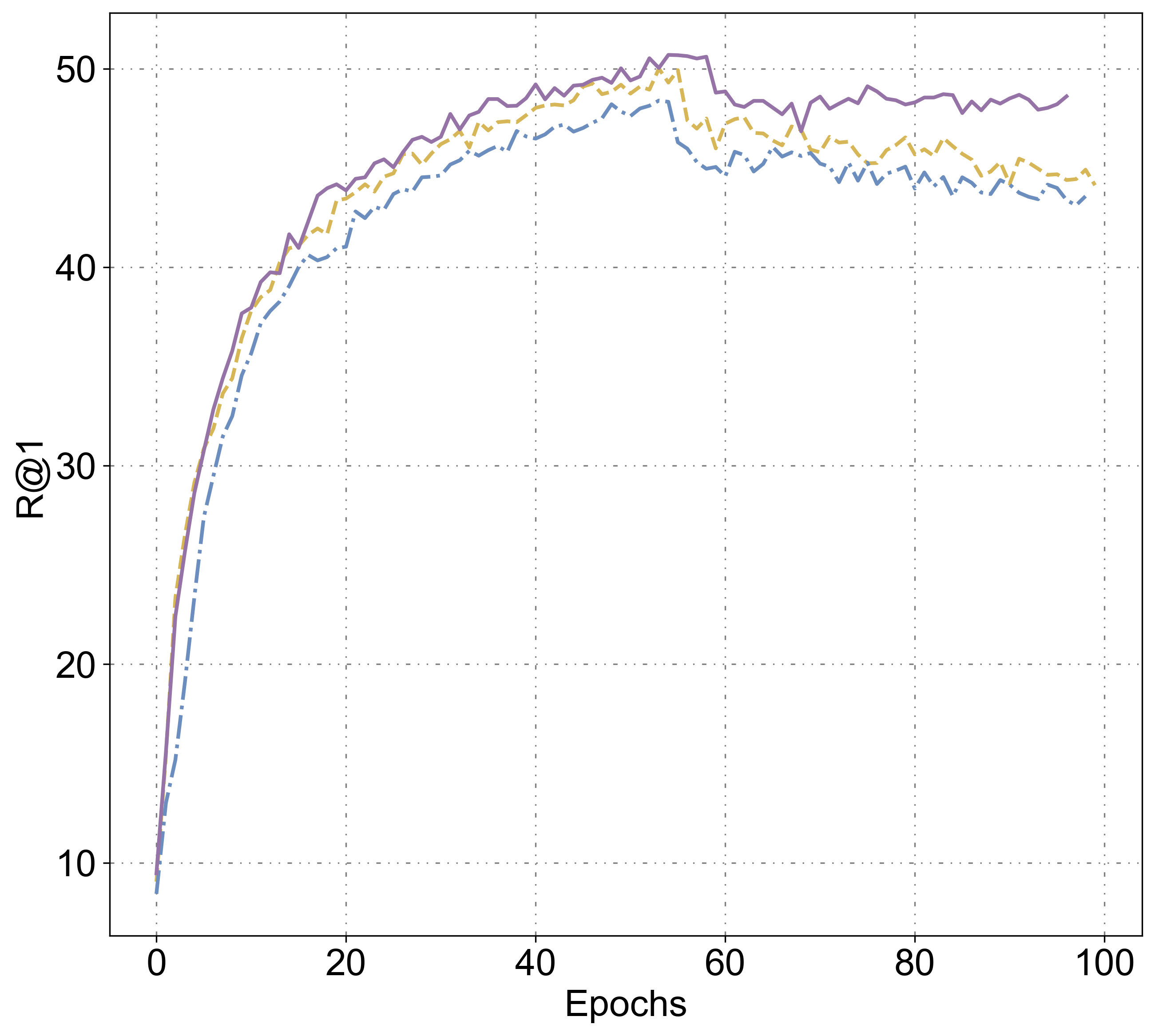}}%
}{2.5em}{4.5em}
\captionof{figure}{Comparison of our proposed GeoAdapter module with different variants of architectural design. With a top-1 recall rate of 50.7 (50.69), our proposed architecture best suits CVGL.}
    \label{fig:compare_geoadapter}
\end{minipage}\hfill\begin{minipage}[t]{.48\linewidth}
    \bottominset{%
    \resizebox{10em}{!}
    {\begin{tabular}{c|c|c|c}
    \hline
       \rowcolor{white!10} No. of $I_L^a$ & R@1 $(\uparrow)$ & R@5 $(\uparrow)$ & R@10 $(\uparrow)$ \\
        \hline
        \rowcolor{lightyellow!100}\textcolor{black}{1} &  \textcolor{black}{52.59} &  \textcolor{black}{59.71} & \textcolor{black}{61.29} \\
        \hline
        \rowcolor{lightblue!100}\textcolor{black}{5} &  \textcolor{black}{{57.37}} & \textcolor{black}{{71.25}} & \textcolor{black}{75.91}\\
        \hline
        \rowcolor{lightviolet!100}\textcolor{black}{10} & \textcolor{black}{54.64} & \textcolor{black}{70.45} & \textcolor{black}{{76.36}}\\
        \hline
    \end{tabular}
    }
    }{%
    \scalebox{1.}{\includegraphics[width=0.9\columnwidth]{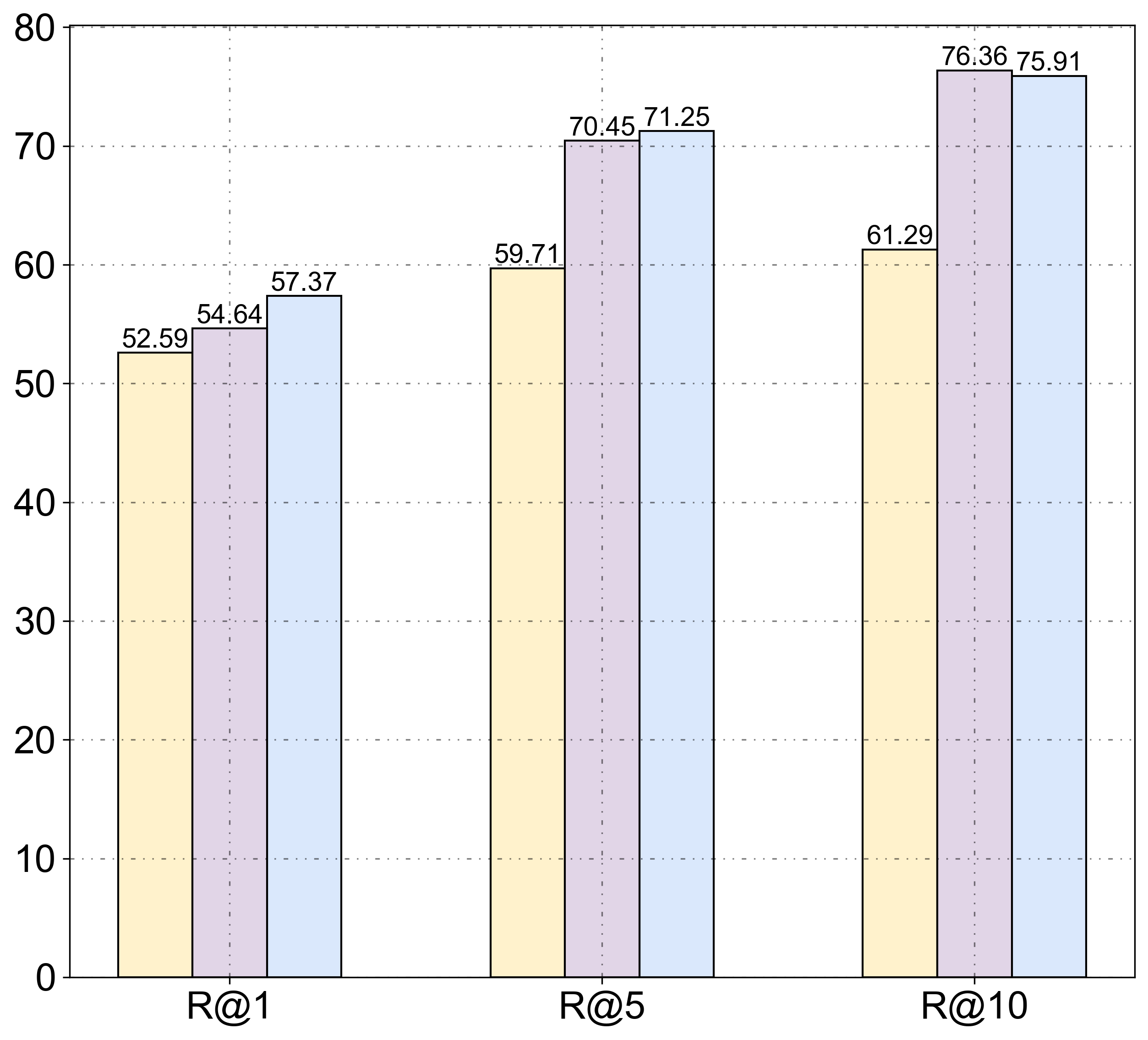}}%
    }{1.5em}{3.5em}
    \centering
    \captionof{figure}{Top-$k$ retrieval recall score of frame-by-frame inference using our method when multiple large aerial images are taken to create the gallery.}
    \label{tab:3}
\end{minipage}
\vspace{-0.3in}
\end{figure}

\vspace{-0.05in}
\label{sec:add_expr}
In this section, we present additional analysis experiments to provide further support for our proposed architecture.\\ 
\textbf{Variants of $\mathbf{G_A}$:} Following the discussion in~\cref{sec:geoadapter}, we compare our proposed GeoAdapter architecture with two other variants. Precisely, in $\mathbf{G_A^\text{ALL}}$, we compute temporal attention across every patch of every input frame. While in $\mathbf{G_A^\text{ASYM}}$, we compute temporal attention across \texttt{[CLS]} tokens in the aerial branch but compute temporal attention across all the patches of every input frame in the street-view branch. In ~\cref{fig:compare_geoadapter}, we compare both these variants with our proposed $\mathbf{G_A}$ where we compute temporal attention across the \texttt{[CLS]} tokens in both branches. It can be inferred from the figure that out of all three variants, the \texttt{[CLS]} token-based temporal attention is best suited for CVGL.\\
\textbf{Effect of Top-$k$ large aerial images:} To further understand the effect of different numbers of large aerial images used to create a small aerial gallery $\mathcal{G}$.~\cref{tab:3} shows the top-$k$ recall rate of frame-by-frame inference using our method when multiple large aerial images are taken to create the gallery.\\
\textbf{Comparison with other Adapters:} In order to showcase the efficacy of our proposed GeoAdapter module $\mathbf{G_A}$, we compare the Sequence-to-Image inference performance of our methodology using adapters from~\cite{yang2022aim} and~\cite{pan2022st}. Specifically, we replace our adapter module with AIM and ST-Adapter without modifying any other component.~\cref{tab:comp_adapt} shows the top-$k$ retrieval recall score for Sequence-to-Image inference with different adapters. It should be noted that the AIM adapter performs comparably to $\mathbf{G_A^\text{ALL}}$ variant due to its similarity in temporal attention computation. In addition to better recall rate, the temporal attention computation in our method can be computationally lighter than in AIM or other variants of $\mathbf{G_A}$ because we only compute attention across \texttt{[CLS]} token of each input frame instead of all input tokens. Additional experiments showing our method's recall rate with varying distance thresholds and runtime analysis are presented in the supplementary material.

\vspace{-0.1in}
\section{Conclusion}
\vspace{-0.1in}

In conclusion, this work introduces \textbf{GAReT}, a fully transformer-based method for cross-view video geo-localization (CVGL) that, unlike prior works, doesn't rely on camera parameters or odometry.
We proposed \textit{GeoAdapter}, a transformer-adapter to aggregate image-level representations of an image geo-localization method and adapt it to video inputs, making our method computationally lightweight than existing works.
% while outperforming them in several benchmark datasets. 
Additionally, to ensure temporally consistent GPS predictions, we proposed \textit{TransRetriever}, an encoder-decoder transformer model that encodes the top-$k$ nearest neighbor predictions for each frame and auto-regressively decodes the best result based on the previous frame's predictions.
% Additionally, we established the issue of temporal inconsistency in GPS predictions in prior works due to modeling per-frame localization as independent nearest-neighbor retrieval operations. Addressing this, we introduced TransRetriever, the first transformer-based model that models independent frame retrievals through an auto-regressive decoder. 
Extensive experimental evaluations demonstrate the effectiveness of our approach, showcasing state-of-the-art performance on benchmark datasets. 
% to ensure temporal consistency in frame-to-frame predictions, we proposed an encoder-decoder transformer model (\textit{TransRetriever}) that encodes the top-$k$ nearest neighbor predictions for each frame and auto-regressively decodes the best result based on previous frames.
\section*{Acknowledgments}
This research is funded by an academic grant from the National Geospatial-Intelligence Agency (NGA) (Award \# HM0476-20-1-0001, Project Title: Estimating Geospatial Trajectories of Videos Using Cross-View Image Matching)
We thank all anonymous reviewers and ACs for their constructive suggestions.
% \clearpage  % TODO REVIEW/FINAL: This \clearpage needs to be removed from both review and camera-ready versions.

\bibliographystyle{splncs04}
\bibliography{egbib}

\begin{thebibliography}{10}
\providecommand{\url}[1]{\texttt{#1}}
\providecommand{\urlprefix}{URL }
\providecommand{\doi}[1]{https://doi.org/#1}

\bibitem{ba2016layer}
Ba, J.L., Kiros, J.R., Hinton, G.E.: Layer normalization. arXiv preprint arXiv:1607.06450  (2016)

\bibitem{berrabah2011visual}
Berrabah, S.A., Sahli, H., Baudoin, Y.: Visual-based simultaneous localization and mapping and global positioning system correction for geo-localization of a mobile robot. Measurement Science and Technology  \textbf{22}(12),  124003 (2011)

\bibitem{bertasius2021space}
Bertasius, G., Wang, H., Torresani, L.: Is space-time attention all you need for video understanding? In: ICML. vol.~2, p.~4 (2021)

\bibitem{berton2022rethinking}
Berton, G., Masone, C., Caputo, B.: Rethinking visual geo-localization for large-scale applications. In: Proceedings of the IEEE/CVF Conference on Computer Vision and Pattern Recognition. pp. 4878--4888 (2022)

\bibitem{brejcha2017state}
Brejcha, J., {\v{C}}ad{\'\i}k, M.: State-of-the-art in visual geo-localization. Pattern Analysis and Applications  \textbf{20},  613--637 (2017)

\bibitem{bresson2021transformer}
Bresson, X., Laurent, T.: The transformer network for the traveling salesman problem. arXiv preprint arXiv:2103.03012  (2021)

\bibitem{brosh2019accurate}
Brosh, E., Friedmann, M., Kadar, I., Yitzhak~Lavy, L., Levi, E., Rippa, S., Lempert, Y., Fernandez-Ruiz, B., Herzig, R., Darrell, T.: Accurate visual localization for automotive applications. In: Proceedings of the IEEE/CVF Conference on Computer Vision and Pattern Recognition Workshops. pp.~0--0 (2019)

\bibitem{cho2014properties}
Cho, K., Van~Merri{\"e}nboer, B., Bahdanau, D., Bengio, Y.: On the properties of neural machine translation: Encoder-decoder approaches. arXiv preprint arXiv:1409.1259  (2014)

\bibitem{deng2009imagenet}
Deng, J., Dong, W., Socher, R., Li, L.J., Li, K., Fei-Fei, L.: Imagenet: A large-scale hierarchical image database. In: 2009 IEEE conference on computer vision and pattern recognition. pp. 248--255. Ieee (2009)

\bibitem{dosovitskiy2020image}
Dosovitskiy, A., Beyer, L., Kolesnikov, A., Weissenborn, D., Zhai, X., Unterthiner, T., Dehghani, M., Minderer, M., Heigold, G., Gelly, S., et~al.: An image is worth 16x16 words: Transformers for image recognition at scale. arXiv preprint arXiv:2010.11929  (2020)

\bibitem{fervers2023uncertainty}
Fervers, F., Bullinger, S., Bodensteiner, C., Arens, M., Stiefelhagen, R.: Uncertainty-aware vision-based metric cross-view geolocalization. In: Proceedings of the IEEE/CVF Conference on Computer Vision and Pattern Recognition. pp. 21621--21631 (2023)

\bibitem{hu2018cvm}
Hu, S., Feng, M., Nguyen, R.M., Lee, G.H.: Cvm-net: Cross-view matching network for image-based ground-to-aerial geo-localization. In: Proceedings of the IEEE Conference on Computer Vision and Pattern Recognition. pp. 7258--7267 (2018)

\bibitem{ioffe2015batch}
Ioffe, S., Szegedy, C.: Batch normalization: Accelerating deep network training by reducing internal covariate shift. In: International conference on machine learning. pp. 448--456. pmlr (2015)

\bibitem{kay2017kinetics}
Kay, W., Carreira, J., Simonyan, K., Zhang, B., Hillier, C., Vijayanarasimhan, S., Viola, F., Green, T., Back, T., Natsev, P., et~al.: The kinetics human action video dataset. arXiv preprint arXiv:1705.06950  (2017)

\bibitem{kingma2014adam}
Kingma, D.P., Ba, J.: Adam: A method for stochastic optimization. arXiv preprint arXiv:1412.6980  (2014)

\bibitem{lentsch2023slicematch}
Lentsch, T., Xia, Z., Caesar, H., Kooij, J.F.: Slicematch: Geometry-guided aggregation for cross-view pose estimation. In: Proceedings of the IEEE/CVF Conference on Computer Vision and Pattern Recognition. pp. 17225--17234 (2023)

\bibitem{liu2019lending}
Liu, L., Li, H.: Lending orientation to neural networks for cross-view geo-localization. In: Proceedings of the IEEE/CVF conference on computer vision and pattern recognition. pp. 5624--5633 (2019)

\bibitem{liu2022video}
Liu, Z., Ning, J., Cao, Y., Wei, Y., Zhang, Z., Lin, S., Hu, H.: Video swin transformer. In: Proceedings of the IEEE/CVF conference on computer vision and pattern recognition. pp. 3202--3211 (2022)

\bibitem{gtsp}
Noon, C., Bean, J.: An efficient transformation of the generalized traveling salesman problem. INFOR. Information Systems and Operational Research  \textbf{31} (02 1993). \doi{10.1080/03155986.1993.11732212}

\bibitem{pan2022st}
Pan, J., Lin, Z., Zhu, X., Shao, J., Li, H.: St-adapter: Parameter-efficient image-to-video transfer learning. Advances in Neural Information Processing Systems  \textbf{35},  26462--26477 (2022)

\bibitem{paszke2019pytorch}
Paszke, A., Gross, S., Massa, F., Lerer, A., Bradbury, J., Chanan, G., Killeen, T., Lin, Z., Gimelshein, N., Antiga, L., et~al.: Pytorch: An imperative style, high-performance deep learning library. Advances in neural information processing systems  \textbf{32} (2019)

\bibitem{radford2021learning}
Radford, A., Kim, J.W., Hallacy, C., Ramesh, A., Goh, G., Agarwal, S., Sastry, G., Askell, A., Mishkin, P., Clark, J., et~al.: Learning transferable visual models from natural language supervision. In: International conference on machine learning. pp. 8748--8763. PMLR (2021)

\bibitem{regmi2021video}
Regmi, K., Shah, M.: Video geo-localization employing geo-temporal feature learning and gps trajectory smoothing. In: Proceedings of the IEEE/CVF International Conference on Computer Vision. pp. 12126--12135 (2021)

\bibitem{beyond}
Shi, Y., Li, H.: Beyond cross-view image retrieval: Highly accurate vehicle localization using satellite image. In: Proceedings of the IEEE/CVF Conference on Computer Vision and Pattern Recognition. pp. 17010--17020 (2022)

\bibitem{shi2019spatial}
Shi, Y., Liu, L., Yu, X., Li, H.: Spatial-aware feature aggregation for image based cross-view geo-localization. Advances in Neural Information Processing Systems  \textbf{32} (2019)

\bibitem{shi2020looking}
Shi, Y., Yu, X., Campbell, D., Li, H.: Where am i looking at? joint location and orientation estimation by cross-view matching. In: Proceedings of the IEEE/CVF Conference on Computer Vision and Pattern Recognition. pp. 4064--4072 (2020)

\bibitem{shi2022accurate}
Shi, Y., Yu, X., Liu, L., Campbell, D., Koniusz, P., Li, H.: Accurate 3-dof camera geo-localization via ground-to-satellite image matching. IEEE Transactions on Pattern Analysis and Machine Intelligence  \textbf{45}(3),  2682--2697 (2022)

\bibitem{shi2022cvlnet}
Shi, Y., Yu, X., Wang, S., Li, H.: Cvlnet: Cross-view semantic correspondence learning for video-based camera localization. In: Asian Conference on Computer Vision. pp. 123--141. Springer (2022)

\bibitem{tian2017cross}
Tian, Y., Chen, C., Shah, M.: Cross-view image matching for geo-localization in urban environments. In: Proceedings of the IEEE Conference on Computer Vision and Pattern Recognition. pp. 3608--3616 (2017)

\bibitem{pmlr-v139-touvron21a}
Touvron, H., Cord, M., Douze, M., Massa, F., Sablayrolles, A., Jegou, H.: Training data-efficient image transformers \&amp; distillation through attention. In: Meila, M., Zhang, T. (eds.) Proceedings of the 38th International Conference on Machine Learning. Proceedings of Machine Learning Research, vol.~139, pp. 10347--10357. PMLR (18--24 Jul 2021), \url{https://proceedings.mlr.press/v139/touvron21a.html}

\bibitem{vaswani2017attention}
Vaswani, A., Shazeer, N., Parmar, N., Uszkoreit, J., Jones, L., Gomez, A.N., Kaiser, {\L}., Polosukhin, I.: Attention is all you need. Advances in neural information processing systems  \textbf{30} (2017)

\bibitem{vyas2022gama}
Vyas, S., Chen, C., Shah, M.: Gama: Cross-view video geo-localization. In: European Conference on Computer Vision. pp. 440--456. Springer (2022)

\bibitem{xia2023convolutional}
Xia, Z., Booij, O., Kooij, J.F.: Convolutional cross-view pose estimation. arXiv preprint arXiv:2303.05915  (2023)

\bibitem{xia2022visual}
Xia, Z., Booij, O., Manfredi, M., Kooij, J.F.: Visual cross-view metric localization with dense uncertainty estimates. In: European Conference on Computer Vision. pp. 90--106. Springer (2022)

\bibitem{yang2021cross}
Yang, H., Lu, X., Zhu, Y.: Cross-view geo-localization with layer-to-layer transformer. Advances in Neural Information Processing Systems  \textbf{34},  29009--29020 (2021)

\bibitem{yang2022aim}
Yang, T., Zhu, Y., Xie, Y., Zhang, A., Chen, C., Li, M.: Aim: Adapting image models for efficient video action recognition. In: The Eleventh International Conference on Learning Representations (2022)

\bibitem{yu2020bdd100k}
Yu, F., Chen, H., Wang, X., Xian, W., Chen, Y., Liu, F., Madhavan, V., Darrell, T.: Bdd100k: A diverse driving dataset for heterogeneous multitask learning. In: Proceedings of the IEEE/CVF conference on computer vision and pattern recognition. pp. 2636--2645 (2020)

\bibitem{zamir2014visual}
Zamir, A.R.: Visual geo-localization and location-aware image understanding  (2014)

\bibitem{zamir2014image}
Zamir, A.R., Shah, M.: Image geo-localization based on multiplenearest neighbor feature matching usinggeneralized graphs. IEEE transactions on pattern analysis and machine intelligence  \textbf{36}(8),  1546--1558 (2014)

\bibitem{zhang2023geodtr+}
Zhang, X., Li, X., Sultani, W., Chen, C., Wshah, S.: Geodtr+: Toward generic cross-view geolocalization via geometric disentanglement. arXiv preprint arXiv:2308.09624  (2023)

\bibitem{geodetr}
Zhang, X., Li, X., Sultani, W., Zhou, Y., Wshah, S.: Cross-view geo-localization via learning disentangled geometric layout correspondence. In: Proceedings of the AAAI Conference on Artificial Intelligence. vol.~37, pp. 3480--3488 (2023)

\bibitem{geometric_layout}
Zhang, X., Li, X., Sultani, W., Zhou, Y., Wshah, S.: Cross-view geo-localization via learning disentangled geometric layout correspondence. In: Proceedings of the AAAI Conference on Artificial Intelligence. vol.~37, pp. 3480--3488 (2023)

\bibitem{zhang2023cross}
Zhang, X., Sultani, W., Wshah, S.: Cross-view image sequence geo-localization. In: Proceedings of the IEEE/CVF Winter Conference on Applications of Computer Vision. pp. 2914--2923 (2023)

\bibitem{zhu2022transgeo}
Zhu, S., Shah, M., Chen, C.: Transgeo: Transformer is all you need for cross-view image geo-localization. In: Proceedings of the IEEE/CVF Conference on Computer Vision and Pattern Recognition. pp. 1162--1171 (2022)

\bibitem{zhu2021revisiting}
Zhu, S., Yang, T., Chen, C.: Revisiting street-to-aerial view image geo-localization and orientation estimation. In: Proceedings of the IEEE/CVF Winter Conference on Applications of Computer Vision. pp. 756--765 (2021)

\bibitem{zhu2021vigor}
Zhu, S., Yang, T., Chen, C.: Vigor: Cross-view image geo-localization beyond one-to-one retrieval. In: Proceedings of the IEEE/CVF Conference on Computer Vision and Pattern Recognition. pp. 3640--3649 (2021)

\end{thebibliography}
\end{document}